\documentclass[sigconf,authorversion]{acmart}

\usepackage[utf8]{inputenc}
\usepackage{xcolor}
\usepackage{colortbl}
\usepackage{hyperref}
\hypersetup{hidelinks}
\usepackage{booktabs}
\usepackage{pifont}%
\usepackage{balance}
\newcommand{\cmark}{\ding{51}}%
\newcommand{\xmark}{\ding{55}}%

\newcommand{\cn}{\textsc{MultiMediate\,}}

\AtBeginDocument{%
  \providecommand\BibTeX{{%
    \normalfont B\kern-0.5em{\scshape i\kern-0.25em b}\kern-0.8em\TeX}}}

\copyrightyear{2023} 
\acmYear{2023} 
\setcopyright{acmlicensed}\acmConference[MM '23]{Proceedings of the 31st ACM International Conference on Multimedia}{October 29-November 3, 2023}{Ottawa, ON, Canada}
\acmBooktitle{Proceedings of the 31st ACM International Conference on Multimedia (MM '23), October 29-November 3, 2023, Ottawa, ON, Canada}
\acmPrice{15.00}
\acmDOI{10.1145/3581783.3613851}
\acmISBN{979-8-4007-0108-5/23/10}

\newcommand\philipp[1]{\textcolor{green}{}}
\newcommand\dominike[1]{\textcolor{orange}{}}
\newcommand\dominik[1]{\textcolor{red}{}}
\newcommand\michal[1]{\textcolor{blue}{}}

\begin{document}

\author{Philipp M\"uller}\thanks{$^*$These authors contributed equally to this work}
\affiliation{
    \institution{DFKI}
    \city{Saarbr\"ucken}
    \country{Germany}
}
\email{philipp.mueller@dfki.de}

\author{Michal Balazia$^{*}$}%
\affiliation{
    \institution{INRIA Sophia Antipolis}
    \city{Sophia Antipolis}
    \country{France}
}
\email{michal.balazia@inria.fr}

\author{Tobias Baur$^{*}$}%
\affiliation{
    \institution{University of Augsburg}
    \city{Augsburg}
    \country{Germany}
}
\email{tobias.baur@uni-a.de}

\author{Michael Dietz$^{*}$}%
\affiliation{
    \institution{University of Augsburg}
    \city{Augsburg}
    \country{Germany}
}
\email{michael.dietz@uni-a.de}

\author{Alexander Heimerl$^{*}$}%
\affiliation{
    \institution{University of Augsburg}
    \city{Augsburg}
    \country{Germany}
}
\email{alexander.heimerl@uni-a.de}

\author{Dominik Schiller$^{*}$}%
\affiliation{
    \institution{University of Augsburg}
    \city{Augsburg}
    \country{Germany}
}
\email{dominik.schiller@uni-a.de}

\author{Mohammed Guermal}%
\affiliation{
    \institution{INRIA Sophia Antipolis}
    \city{Sophia Antipolis}
    \country{France}
}
\email{mohammed.guermal@inria.fr}

\author{Dominike Thomas}%
\affiliation{
    \institution{University of Stuttgart}
    \city{Stuttgart}
    \country{Germany}
}
\email{dominike.thomas@gmail.com}

\author{Fran\c{c}ois Br\'emond}
\affiliation{
    \institution{INRIA Sophia Antipolis}
    \city{Sophia Antipolis}
    \country{France}
}
\email{francois.bremond@inria.fr}

\author{Jan Alexandersson}
\affiliation{
    \institution{DFKI}
    \city{Saarbr\"ucken}
    \country{Germany}
}
\email{janal@dfki.de}

\author{Elisabeth Andr\'e}
\affiliation{
    \institution{University of Augsburg}
    \city{Augsburg}
    \country{Germany}
}
\email{elisabeth.andre@uni-a.de}

\author{Andreas Bulling}
\affiliation{
    \institution{University of Stuttgart}
    \city{Stuttgart}
    \country{Germany}
}
\email{andreas.bulling@vis.uni-stuttgart.de}

\title{\cn'23: Engagement Estimation and Bodily Behaviour Recognition in Social Interactions}
\date{May 1, 2023}

\renewcommand{\shortauthors}{Philipp Müller et al.}

\begin{abstract}

Automatic analysis of human behaviour is a fundamental prerequisite for the creation of machines that can effectively interact with- and support humans in social interactions.
In \cn'23, we address two key human social behaviour analysis tasks for the first time in a controlled challenge: engagement estimation and bodily behaviour recognition in social interactions.
This paper describes the \cn'23 challenge and presents novel sets of annotations for both tasks.
For engagement estimation we collected novel annotations on the NOvice eXpert Interaction (NOXI) database.
For bodily behaviour recognition, we annotated test recordings of the MPIIGroupInteraction corpus with the BBSI annotation scheme.
In addition, we present baseline results for both challenge tasks.

\end{abstract}



\maketitle

\section{Introduction}

Artificial mediators~\cite{park2020investigating}, i.e. interactive intelligent agents that actively engage in a conversation in a human-like way have the potential to positively influence the course and/or outcomes of human interactions.
They have been studied in a variety of contexts, including collaborative teamwork~\cite{bohus2010facilitating,short2017robot}, mental health~\cite{birmingham2020can}, and education~\cite{lopes2017first,engwall2020interaction}. 
A central prerequisite for effective and context-aware artificial mediation is the ability to comprehensively detect- and interpret the diverse set of social signals expressed by humans.
At present, this challenge is still largely unsolved, and research on artificial mediators often has to rely on Wizard-of-Oz paradigms~\cite{lopes2017first,utami2019collaborative,engwall2020interaction,birmingham2020can,ohshima_neut_2017,sebo2020robots}.

With the multi-year \cn challenge we contribute to realising the vision of autonomous artificial mediators by facilitating measurable advances on central conversational behaviour sensing and analysis tasks.
The first iteration of the challenge in 2021~\cite{muller2021multimediate} has addressed eye contact detection and next speaker prediction while \cn'22 has focused on backchannel analysis~\cite{muller2022multimediate,amer2023backchannel}.
In two separate tracks, \cn'23 addresses the recognition of complex bodily behaviours, as well as the estimation of a persons' engagement level.
Bodily behaviours such as fumbling, folded arms, or gesturing are a key social signal and were shown to be connected to many important high-level phenomena including stress regulation, attraction, or social verticality~\cite{carney2005,vacharkulksemsuk2016dominant,hall2005,mohiyeddini2013displacement}.
As a result, accurate recognition of bodily behaviours can serve as a building block for the recognition of such more abstract phenomena.
Knowing how engaged participants are, individually or as a group, is important for a mediator whose goal it is to keep engagement at a high level. 
Engagement is closely linked to the previous \cn tasks of eye contact detection~\cite{oertel_gaze-based_2013, peters_engagement_2005} as well as backchanneling~\cite{goswami_towards_2020}.

With \cn'23 we present the first challenge on engagement estimation and the recognition of bodily behaviours in social interaction.
We define the tasks and evaluation criteria and describe new annotations collected on the NOvice eXpert Interaction (NOXI) database~\cite{Cafaro:2017}, as well as on unreleased test recordings of MPIIGroupInteraction~\cite{muller_detecting_2018}.
Furthermore, we present baseline approaches for both challenge tasks and report evaluation results.
We make all collected annotations, baseline implementations, and raw feature representations publicly available for further use, even beyond the scope of \cn'23.\footnote{\url{https://multimediate-challenge.org}}

\section{Related Work}

We review previous works on methods and datasets for engagement estimation and bodily behaviour recognition in social interaction.

\subsection{Engagement Estimation}

Engagement has been investigated from various research angles, e.g. how to define, annotate, or to automatically predict it. Rich et al. \cite{CharlesRichEngagementHumanRobotInteraction} introduced a module for the recognition of engagement in human-robot interaction based on backchannels. Sanghvi et al. \cite{Sanghvi:2011:AAA:1957656.1957781BodyMotion} predicted engagement based on body posture features. Bednarik et al. \cite{Bednarik:2012:GCE:2401836.2401846Engagement} focused on recognizing conversational engagement with gaze data.
Research in detecting engagement in students is prolific and promising~\citep{karimah_automatic_2021, goldberg_attentive_2021}. Engagement is also often studied in children~\citep{rajagopalan_play_2015} and, more particularly, in children interacting with an artificial agent~\citep{oertel_engagement_2020, park_model-free_2019, jain_modeling_2020}. \citet{guhan_met_2022} researched engagement in mental health patients, based on videos of the patient. 
Some datasets also offer engagement ratings, such as RECOLA~\cite{ringeval_introducing_2013}, MHHRI~\cite{celiktutan_multimodal_2019}, and~\cite{hradis_voice_2012} with annotations from~\cite{Bednarik:2012:GCE:2401836.2401846Engagement}.
In \autoref{tab:datasets} we provide an overview over the existing social interaction datasets with engagement annotations.
The NoXi dataset annotated for \cn'23 is significantly larger compared to previous datasets.

\subsection{Bodily Behaviour Recognition}

Bodily behaviours are key signals in social interactions and are related to many higher-level attributes.
For example, displacement behaviours (e.g. fumbling, face-touching, or grooming) are associated with anxiety and stress regulation~\cite{bardi2011behavioral,mohiyeddini2015neuroticism,mohiyeddini2013displacement}.
Leaning towards the interlocutor is connected with rapport~\cite{sharpley1995} and crossed arms can be indicative of emotion expressions~\cite{wallbott1998}.
Further connections were found between bodily behaviours and liking~\cite{mehrabian1968relationship,mehrabian1969encoding}, attractiveness~\cite{vacharkulksemsuk2016dominant}, and social verticality~\cite{hall2005}.

Despite this importance, little previous work addressed the recognition of bodily behaviours like fumbling, grooming, crossed arms, or gesturing in social interactions~\cite{balazia2022bodily,liu2021imigue}.
While impressive progress was made on body- and hand pose estimation~\cite{cao2017realtime,simon2017hand}, it is not a trivial task to establish the connection between low-level keypoint detections and complex bodily behaviours that are relevant to the interaction.
Furthermore, only a limited number of bodily behaviour recognition datasets containing spontaneous behaviour in social interactions is available.
The PAVIS Face-Touching dataset~\cite{beyan2020analysis} consists of a single annotated behaviour (face touching) in group discussions.
The iMiGUE dataset~\cite{liu2021imigue} contains annotations of 32 behaviour classes annotated for speakers at sports press conferences.
For the purpose of \cn, the recently published BBSI dataset~\cite{balazia2022bodily} is most relevant, which consists of 15 behaviour classes annotated for all participants of 3-4 person group conversations.
Such group conversations are one of the main application domains of artificial mediators.

\begin{table}[t]
\begin{tabular}{l|cccc}
    Corpus & Screen & Group size & Length & Part. \\
    \midrule 
    \citet{guhan_met_2022} & \cmark  & 2 & 1h5m & 13\\
    RECOLA~\cite{ringeval_introducing_2013} & \cmark & 2 & 3h50m & 46 \\
    \citet{Bednarik:2012:GCE:2401836.2401846Engagement} & \cmark & 4-7 & 6h & 9 groups \\
    MMHRI~\cite{celiktutan_multimodal_2019} & \xmark & 2 & 6h & 18 \\
    \midrule
    NOXI (ours) & \cmark & 2  & 25h & 87\\
\end{tabular}
\caption{Social interaction datasets with engagement annotations, excluding MOOC and school settings and children as participants. 
\textit{Screen} indicates whether interaction was screen-mediated, \textit{Group size} the number of humans per interaction, \textit{Length} the total duration of interactions, and \textit{Part.} the total number of human participants.
}
\label{tab:datasets}
\end{table}

\section{Challenge Description}

In the following we present the two challenge tasks and the utilised datasets.
For both tasks test samples (without ground truth) are released to participants before the challenge deadline.
Participants in turn submit their predictions for evaluation.

\subsection{Engagement Estimation Task}

\paragraph{Task definition}

The task includes the continuous, frame-wise prediction of the level of conversational engagement of each participant on a continuous scale from 0 (lowest) to 1 (highest). Participants are encouraged to investigate multimodal as well as reciprocal behaviour of both interlocutors in the Novice-Expert Interaction corpus. We make use of the Concordance Correlation Coefficient (CCC)~\cite{ccc} to evaluate predictions on the test set.

\begin{figure}[h]
\centering
\includegraphics[width=1.0\columnwidth]{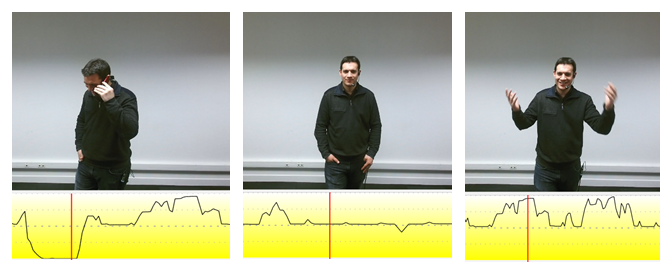}
\caption{Snapshots of scenes of a participant in the NOXI corpus being disengaged (left), neutral (center) and highly engaged (right).}
\label{fig:noxi:setup-5}
\end{figure}

\paragraph{Dataset}

The NOvice eXpert Interaction (\textsc{NOXI}) database \cite{Cafaro:2017} is a corpus of dyadic, screen-mediated face-to-face interactions in an expert-novice knowledge sharing context. In a session, one participant assumes the role of an expert and the other participant the role of a novice. 
Figure \ref{fig:noxi:setup-5} shows two users during interaction.
NOXI includes interactions recorded at three locations (France, Germany and UK), spoken in eight languages (English, French, German, Spanish, Indonesian, Arabic, Dutch and Italian), 
discussing a wide range of topics. The dataset offers over 25 hours (x2) of recordings of dyadic interactions in natural settings, featuring synchronized audio, video (25fps), and motion capture data (using a Kinect 2.0). 
We will use subset of this corpus containing 48 sessions for training and 16 sessions for testing (75/25 split). We aimed to obtain data of spontaneous behavior in a natural setting on a variety of discussion topics. Therefore, one of the main design goals was to match recorded participants based on their common interests. This means that we first gathered
potential experts willing to share their knowledge about one or more topics they were knowledgeable and passionate about, and secondly we recruited novices willing to discuss or learn more about the available set of topics offered by experts.
The corpus further introduces interruptions of the novices in order to provoke experts' reactions when conversational engagement gets interrupted.
In particular, for this challenge, each session has been annotated in a continuous matter, meaning each video frame has a score between 0 and 1. Each rating was performed by at least two (up to 7) annotators (Average: 3.6 raters per session). We created gold standard annotations by calculating the mean over all raters. 
The NOXI dataset can be obtained from the website\footnote{\label{dataset-link}\url{https://multimediate-challenge.org/datasets/Dataset_NoXi/}}.

\begin{figure}[t]
  \centering
  \includegraphics[width=\columnwidth]{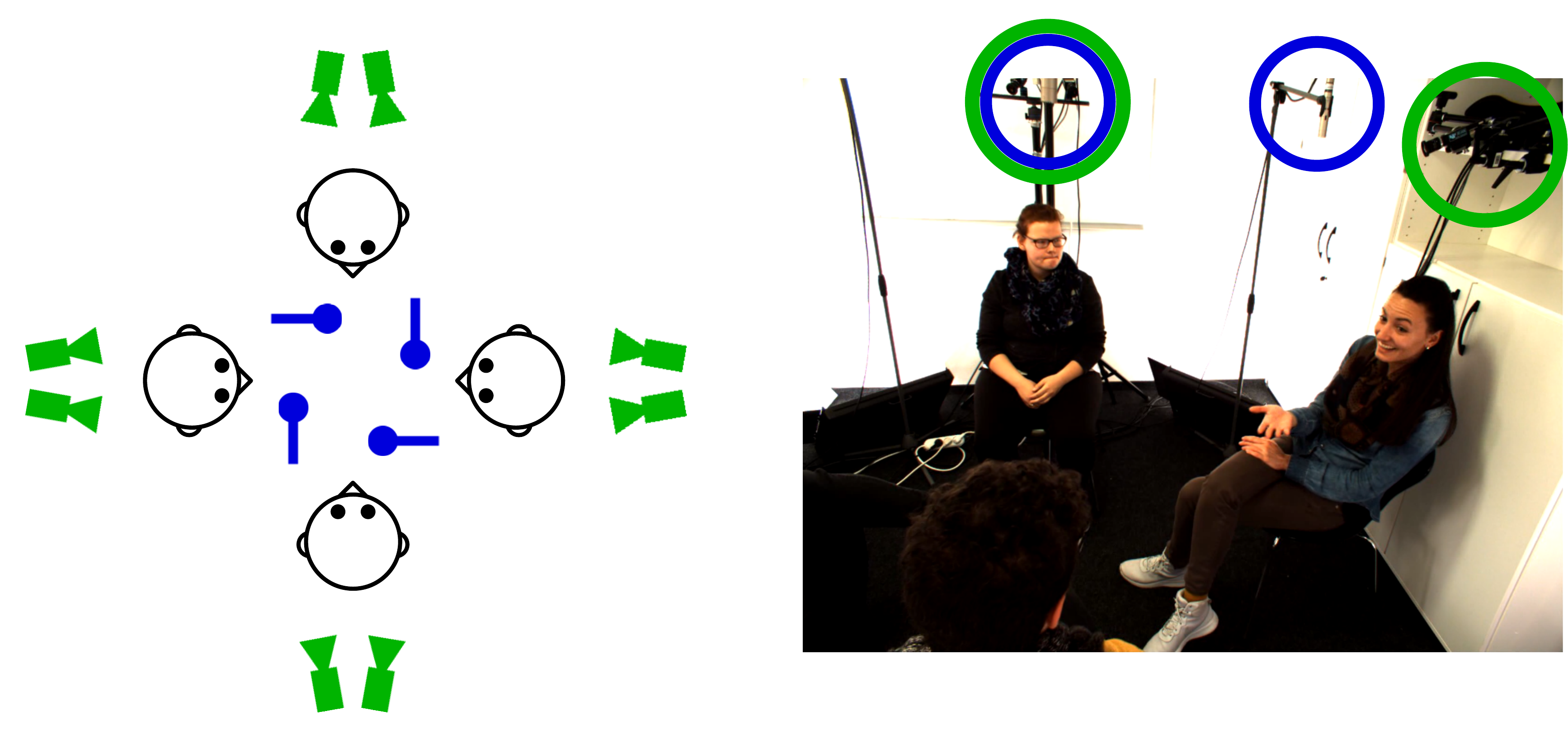}
  \caption{Setup of the MPIIGroupInteraction dataset. %
  Reproduced with permission from the authors of~\cite{muller_detecting_2018}.
  }~\label{fig:study_design_iui18}
\end{figure}

\subsection{Bodily Behaviour Recognition Task}

\paragraph{Task definition}
We formulate bodily behaviour recognition as a multi-label classification task.
Challenge participants are required to predict which of 15 behaviour classes are present in a 64 (2.13 sec) frame input window.
For each 64-frame window, we provide a frontal view on the target participant, as well as two side views (left and right).
As the behaviour classes on this task are highly unbalanced, we will measure performance using average precision computed per class and aggregated using macro averaging, i.e. giving the same weight to each class.
This encourages challenge competitors to develop novel methods to improve performance on challenging low-frequency classes.

\paragraph{Dataset}
As in \cn'21~\cite{muller2021multimediate}, our challenge is based on the MPIIGroupInteraction dataset~\cite{muller_detecting_2018,muller2018robust}.
This dataset has served as a basis for diverse tasks, including emergent leadership detection~\cite{muller2019emergent}, eye contact detection~\cite{muller2018robust,fu2021using,ma2022ta}, next speaker prediction~\cite{birmingham2021group}, backchannel analysis~\cite{sharma2022graph,amer2023backchannel}, and body language detection~\cite{balazia2022bodily}.
The MPIIGroupInteraction corpus consists of 22 group discussions between three to four people, each lasting for 20 minutes~\cite{muller_detecting_2018}.
This year's bodily behaviour task is based on the recently collected BBSI annotations~\cite{balazia2022bodily}, consisting of 15 bodily behaviour classes annotated on the whole MPIIGroupInteraction corpus.
For \cn'23, we excluded ``Lean towards'' as inter-annotator agreement was reported to be very low on this class.
We collected bodily behaviour annotations for the remaining 14 classes on 996 samples obtained from six unpublished test recordings of MPIIGroupInteraction following the BBSI protocol~\cite{balazia2022bodily}.
To reach high-quality annotations on the test set, we obtained consensus decisions from three annotators. 
All classes except the ``Stretching'' class were present on the test set.
The MPIIGroupInteraction dataset can be obtained from the website\footnote{\label{dataset-link}\url{https://multimediate-challenge.org/datasets/Dataset_MPII/}}.

\section{Experiments and Results}
We are providing a baseline model for each task. 
This section describes the training methodology as well as the utilized features and results achieved for both tasks.

\subsection{Engagement Estimation}

\subsubsection{Approach}
For the engagement estimation task we rely on a set of multimodal features comprising body posture, facial features and vocal features, followed by a fully connected neural network with three hidden layers of size 112 each.
To prevent overfitting we rely on a dropout layer after the second hidden layer with a dropout rate of 0.25. 
The network has been trained using the Adam optimizer and the mean squared error loss function. 
All hyperparameters have been optimized using the hyperband search algorithm of the KerasTuner framework \cite{omalley2019kerastuner}.

\noindent\textit{Head Features.} 
We extracted features from participants' head and face using OpenFace 2.0~\cite{baltrusaitis2018openface}. 
All features where extracted for each video frame.
The resulting feature vectors are consisting of 68 3D facial landmarks, 56 3D eye landmarks, presence and intensity of 18 action units as well as markers for detection success, detection certainty facial position and rotation.
Furthermore, we also use 17 action units provided by the Microsoft Kinect sensor. 

\noindent\textit{Pose Features.} 
We extract body pose estimates using OpenPose~\cite{cao2017realtime} as well as the Microsoft Kinect sensor data, resulting in the estimation of 350 data points comprising information about the location of various joints as well as their rotation.

\noindent\textit{Voice Features.} 
For the paralinguistic assessment of engagement we extracted two feature sets over a one-second sliding window with a stride of 40ms to match the frame rate of the video stream.
The first feature set is the Geneva Minimalistic Acoustic Parameter Set (eGeMAPS)~\cite{eyben2015geneva}.
This set consists of 54 acoustic parameters that are commonly applied to tasks like depression, mood, and emotion recognition~\cite{valstar2016avec}.
Secondly, we used pretrained version of Soundnet \cite{Aytar2016SoundNet} to extract sound embeddings from the raw signal.  
Soundnet is a deep convolutional neural network that has already been shown to provide effective features for vocal social signal analysis  \cite{DBLP:conf/interspeech/0001SSA18}.

In our baseline approach, we fused the feature vectors of all modalities into one feature vector. 
As a large number of features can lead to overfitting we applied a PCA, reducing the number of features to 83 principal components.

\subsubsection{Results}
The results are depicted in Table \ref{tab:results_engagement}. 
Among the single modalities the vocal features are clearly outperforming the body and head features on the validation set as well as on the test test. 
However, the multimodal feature fusion shows that the combination of all features still outperforms just using vocal features substantially. 
The additional value added by head and body features indicates that the expression of engagement is not clearly bound to one modality but should be analyzed considering multiple modalities.

\begin{table}[]
\begin{tabular}{lcc}
\toprule
Features             & Val CCC & Test CCC\\
\midrule
\textit{Head}        &      &           \\
\ \ openface             & 0.23 & 0.21      \\
\ \ AUs                  & 0.31 & 0.22      \\
\midrule
\textit{Body}        &      &           \\
\ \ skeleton             & 0.47 & 0.43      \\
\ \ openpose             & 0.53  & 0.43     \\
\midrule
\textit{Voice}       &       &          \\
\ \ gemaps               & 0.58  & 0.55     \\
\ \ soundnet             & 0.54  & 0.49     \\
\midrule
\textit{Multimodal}           & & \\
\ \ feature fusion + pca & \cellcolor{gray!25}\textbf{0.71}  & \cellcolor{gray!25}\textbf{0.59}\\
\bottomrule
\end{tabular}
\caption{Concordance correlation coefficient (CCC) of our baseline on engagement detection validation and test sets.}
\label{tab:results_engagement}
\end{table}

\subsection{Bodily Behaviour Recognition}

\subsubsection{Approach}

As our baseline solution, we chose the Video Swin Transformer~\cite{liu2021Swin}, which produced recent state-of-the-art results in action recognition tasks.
It operates on fixed inputs of length $32$ frames and size of $224 \times 224$ pixels. Given the input videos of length $64$ frames and of larger resolutions, we set the stride to $2$, that is we took every second frame, and we resized the video accordingly. 
We assigned input clips with multiple corresponding behavior class labels and clips of different viewpoints are treated as independent samples during training. 
To the clips with no labels, we assigned a new behavior class called \textit{Background}, and, instead of the 14, trained the model in a 15-class multi-label setup. 
To aggregate predictions across views at test time, we averaged the scores obtained from all three views.
We used the Swin Base model that is pre-trained on ImageNet and Kinetics-400, and fine-tuned it on the MPIIGroupInteraction dataset for only one epoch with learning rate $10^{-3}$ and with AdamW optimizer. 
Our implementation uses the open-source toolbox MMaction2~\cite{2020mmaction2} built on top of PyCharm.

\subsubsection{Results}
Results of multiple ablations are reported in Table \ref{tab:results_behaviors}.
We evaluated our approach against ablations that operate on single views, against an aggregation strategy using the maximum across views, as well as against not using an additional background class during training.
The best mean average precision (MAP) on both validation and test sets was achieved by averaging across views and training with a background class.
While the inclusion of the background class only led to minor improvements, averaging across views yielded consistent improvements.
The best single view was the frontal view, and side views resulted in a significant performance drop.
All results clearly outperformed the random baseline.
Results on the test set tend are systematically higher, likely as a result of the higher quality annotations, and the lack of the ``Stretching'' class on the test set which as a result is always evaluated with 1.

\begin{table}[]
\begin{tabular}{lccc}
\toprule
Approach & Val MAP & Test MAP\\
\midrule
random baseline & 0.0884 & 0.2355 \\
\midrule

w/o bkgd class, frontal view & 0.3974 & 0.5315 \\
w/o bkgd class, side view 1 & 0.3030 & 0.4341 \\
w/o bkgd class, side view 2 & 0.3628 & 0.4893 \\
w/o bkgd class, max of views & 0.4087 & 0.5333 \\
w/o bkgd class, mean of views & 0.4084 & 0.5402 \\
w/ bkgd class, frontal view & 0.4051 & 0.5498 \\
w/ bkgd class, side view 1 & 0.3096 & 0.4451 \\
w/ bkgd class, side view 2 & 0.3686 & 0.4641 \\
w/ bkgd class, max of views & 0.4062 & 0.5443 \\
w/ bkgd class, mean of views & \cellcolor{gray!25}\textbf{0.4099} & \cellcolor{gray!25}\textbf{0.5628} \\
\bottomrule
\end{tabular} %
\caption{Validation and test results for the random baseline and different variants of the Video Swin Transformer.}
\label{tab:results_behaviors}
\end{table}

\section{Conclusion}
We introduced \cn'23, the first challenge addressing engagement estimation and bodily behaviour recognition in social interactions in well-defined conditions.
We presented publicly available datasets and evaluation protocols for both tasks, and evaluated baseline approaches.
The evaluation server will remain accessible to researchers even beyond the \cn challenge, contributing to continuing progress on both tasks.

\begin{acks}
P. M\"uller and J. Alexandersson were funded by the \grantsponsor{01IS20075}{German Ministry for Education and Research (BMBF)}{https://www.bmbf.de/}, grant number \grantnum{}{01IS20075} and by the European Union Horizon Europe programme, grant number \grantnum{}{101078950}.
A. Bulling was funded by the European Research Council (ERC; grant agreement 801708). 
M. Balazia was funded by the \grantsponsor{}{French National Research Agency}{https://anr.fr/} under the UCA\textsuperscript{JEDI} Investments into the Future, project number \grantnum{}{ANR-15-IDEX-01}.
The researchers from Augsburg University were partially funded by the Deutsche Forschungsgemeinschaft (DFG, German Research Foundation), project Panorama, grant number \grantnum{442607480}{442607480}.
\end{acks}

\newpage
\bibliographystyle{ACM-Reference-Format}
\balance
\bibliography{bibliography}


\begin{thebibliography}{60}


\ifx \showCODEN    \undefined \def \showCODEN     #1{\unskip}     \fi
\ifx \showDOI      \undefined \def \showDOI       #1{#1}\fi
\ifx \showISBNx    \undefined \def \showISBNx     #1{\unskip}     \fi
\ifx \showISBNxiii \undefined \def \showISBNxiii  #1{\unskip}     \fi
\ifx \showISSN     \undefined \def \showISSN      #1{\unskip}     \fi
\ifx \showLCCN     \undefined \def \showLCCN      #1{\unskip}     \fi
\ifx \shownote     \undefined \def \shownote      #1{#1}          \fi
\ifx \showarticletitle \undefined \def \showarticletitle #1{#1}   \fi
\ifx \showURL      \undefined \def \showURL       {\relax}        \fi
\providecommand\bibfield[2]{#2}
\providecommand\bibinfo[2]{#2}
\providecommand\natexlab[1]{#1}
\providecommand\showeprint[2][]{arXiv:#2}

\bibitem[Amer et~al\mbox{.}(2023)]%
        {amer2023backchannel}
\bibfield{author}{\bibinfo{person}{Ahmed Amer}, \bibinfo{person}{Chirag
  Bhuvaneshwara}, \bibinfo{person}{Gowtham~K Addluri},
  \bibinfo{person}{Mohammed~M Shaik}, \bibinfo{person}{Vedant Bonde}, {and}
  \bibinfo{person}{Philipp M{\"u}ller}.} \bibinfo{year}{2023}\natexlab{}.
\newblock \showarticletitle{Backchannel Detection and Agreement Estimation from
  Video with Transformer Networks}.
\newblock \bibinfo{journal}{\emph{arXiv preprint arXiv:2306.01656}}
  (\bibinfo{year}{2023}).
\newblock


\bibitem[Aytar et~al\mbox{.}(2016)]%
        {Aytar2016SoundNet}
\bibfield{author}{\bibinfo{person}{Yusuf Aytar}, \bibinfo{person}{Carl
  Vondrick}, {and} \bibinfo{person}{Antonio Torralba}.}
  \bibinfo{year}{2016}\natexlab{}.
\newblock \showarticletitle{SoundNet: Learning sound representations from
  unlabeled video}. In \bibinfo{booktitle}{\emph{Advances in Neural Information
  Processing Systems}}. \bibinfo{pages}{892--900}.
\newblock


\bibitem[Balazia et~al\mbox{.}(2022)]%
        {balazia2022bodily}
\bibfield{author}{\bibinfo{person}{Michal Balazia}, \bibinfo{person}{Philipp
  M{\"u}ller}, \bibinfo{person}{{\'A}kos~Levente T{\'a}nczos},
  \bibinfo{person}{August~von Liechtenstein}, {and}
  \bibinfo{person}{Fran{\c{c}}ois Br{\'e}mond}.}
  \bibinfo{year}{2022}\natexlab{}.
\newblock \showarticletitle{Bodily behaviors in social interaction: Novel
  annotations and state-of-the-art evaluation}. In
  \bibinfo{booktitle}{\emph{Proc. of the ACM International Conference on
  Multimedia}}. \bibinfo{pages}{70--79}.
\newblock
\urldef\tempurl%
\url{https://doi.org/10.1145/3503161.3548363}
\showDOI{\tempurl}


\bibitem[Baltrusaitis et~al\mbox{.}(2018)]%
        {baltrusaitis2018openface}
\bibfield{author}{\bibinfo{person}{Tadas Baltrusaitis}, \bibinfo{person}{Amir
  Zadeh}, \bibinfo{person}{Yao~Chong Lim}, {and}
  \bibinfo{person}{Louis-Philippe Morency}.} \bibinfo{year}{2018}\natexlab{}.
\newblock \showarticletitle{Openface 2.0: Facial behavior analysis toolkit}. In
  \bibinfo{booktitle}{\emph{Proc. of the IEEE International Conference on
  Automatic Face \& Gesture Recognition}}. IEEE, \bibinfo{pages}{59--66}.
\newblock
\urldef\tempurl%
\url{https://doi.org/10.1109/FG.2018.00019}
\showDOI{\tempurl}


\bibitem[Bardi et~al\mbox{.}(2011)]%
        {bardi2011behavioral}
\bibfield{author}{\bibinfo{person}{M Bardi}, \bibinfo{person}{T Koone},
  \bibinfo{person}{S Mewaldt}, {and} \bibinfo{person}{K O'Connor}.}
  \bibinfo{year}{2011}\natexlab{}.
\newblock \showarticletitle{Behavioral and physiological correlates of stress
  related to examination performance in college chemistry students}.
\newblock \bibinfo{journal}{\emph{Stress}} \bibinfo{volume}{14},
  \bibinfo{number}{5} (\bibinfo{year}{2011}), \bibinfo{pages}{557--566}.
\newblock
\urldef\tempurl%
\url{https://doi.org/10.3109/10253890.2011.571322}
\showDOI{\tempurl}


\bibitem[Bednarik et~al\mbox{.}(2012)]%
        {Bednarik:2012:GCE:2401836.2401846Engagement}
\bibfield{author}{\bibinfo{person}{Roman Bednarik}, \bibinfo{person}{Shahram
  Eivazi}, {and} \bibinfo{person}{Michal Hradis}.}
  \bibinfo{year}{2012}\natexlab{}.
\newblock \showarticletitle{Gaze and Conversational Engagement in Multiparty
  Video Conversation: An Annotation Scheme and Classification of High and Low
  Levels of Engagement}. In \bibinfo{booktitle}{\emph{Proc. of the 4th Workshop
  on Eye Gaze in Intelligent Human Machine Interaction}}.
\newblock
\urldef\tempurl%
\url{https://doi.org/10.1145/2401836.2401846}
\showDOI{\tempurl}


\bibitem[Beyan et~al\mbox{.}(2020)]%
        {beyan2020analysis}
\bibfield{author}{\bibinfo{person}{Cigdem Beyan}, \bibinfo{person}{Matteo
  Bustreo}, \bibinfo{person}{Muhammad Shahid}, \bibinfo{person}{Gian~Luca
  Bailo}, \bibinfo{person}{Nicolo Carissimi}, {and} \bibinfo{person}{Alessio
  Del~Bue}.} \bibinfo{year}{2020}\natexlab{}.
\newblock \showarticletitle{Analysis of face-touching behavior in large scale
  social interaction dataset}. In \bibinfo{booktitle}{\emph{Proc. of the ACM
  International Conference on Multimodal Interaction}}.
  \bibinfo{pages}{24--32}.
\newblock
\urldef\tempurl%
\url{https://doi.org/10.1145/3382507.3418876}
\showDOI{\tempurl}


\bibitem[Birmingham et~al\mbox{.}(2020)]%
        {birmingham2020can}
\bibfield{author}{\bibinfo{person}{Chris Birmingham}, \bibinfo{person}{Zijian
  Hu}, \bibinfo{person}{Kartik Mahajan}, \bibinfo{person}{Eli Reber}, {and}
  \bibinfo{person}{Maja~J. Mataric}.} \bibinfo{year}{2020}\natexlab{}.
\newblock \showarticletitle{Can {I} Trust You? A User Study of Robot Mediation
  of a Support Group}.
\newblock \bibinfo{journal}{\emph{arXiv preprint arXiv:2002.04671}}
  (\bibinfo{year}{2020}).
\newblock


\bibitem[Birmingham et~al\mbox{.}(2021)]%
        {birmingham2021group}
\bibfield{author}{\bibinfo{person}{Chris Birmingham}, \bibinfo{person}{Kalin
  Stefanov}, {and} \bibinfo{person}{Maja~J Mataric}.}
  \bibinfo{year}{2021}\natexlab{}.
\newblock \showarticletitle{Group-Level Focus of Visual Attention for Improved
  Next Speaker Prediction}. In \bibinfo{booktitle}{\emph{Proc. of the ACM
  International Conference on Multimedia}}. \bibinfo{pages}{4838--4842}.
\newblock
\urldef\tempurl%
\url{https://doi.org/10.1145/3474085.3479213}
\showDOI{\tempurl}


\bibitem[Bohus and Horvitz(2010)]%
        {bohus2010facilitating}
\bibfield{author}{\bibinfo{person}{Dan Bohus} {and} \bibinfo{person}{Eric
  Horvitz}.} \bibinfo{year}{2010}\natexlab{}.
\newblock \showarticletitle{Facilitating multiparty dialog with gaze, gesture,
  and speech}. In \bibinfo{booktitle}{\emph{Proc. of the ACM International
  Conference on Multimodal Interfaces and the Workshop on Machine Learning for
  Multimodal Interaction}}. \bibinfo{pages}{1--8}.
\newblock
\urldef\tempurl%
\url{https://doi.org/10.1145/1891903.1891910}
\showDOI{\tempurl}


\bibitem[Cafaro et~al\mbox{.}(2017)]%
        {Cafaro:2017}
\bibfield{author}{\bibinfo{person}{Angelo Cafaro}, \bibinfo{person}{Johannes
  Wagner}, \bibinfo{person}{Tobias Baur}, \bibinfo{person}{Soumia Dermouche},
  \bibinfo{person}{Mercedes Torres~Torres}, \bibinfo{person}{Catherine
  Pelachaud}, \bibinfo{person}{Elisabeth Andr{\'e}}, {and}
  \bibinfo{person}{Michel~F. Valstar}.} \bibinfo{year}{2017}\natexlab{}.
\newblock \showarticletitle{The NoXi Database: Multimodal Recordings of
  Mediated Novice-Expert Interactions}. In \bibinfo{booktitle}{\emph{Proc. of
  the International Conference on Multimodal Interaction}}.
\newblock
\urldef\tempurl%
\url{https://doi.org/10.1145/3136755.3136780}
\showDOI{\tempurl}


\bibitem[Cao et~al\mbox{.}(2017)]%
        {cao2017realtime}
\bibfield{author}{\bibinfo{person}{Zhe Cao}, \bibinfo{person}{Tomas Simon},
  \bibinfo{person}{Shih-En Wei}, {and} \bibinfo{person}{Yaser Sheikh}.}
  \bibinfo{year}{2017}\natexlab{}.
\newblock \showarticletitle{Realtime multi-person 2d pose estimation using part
  affinity fields}. In \bibinfo{booktitle}{\emph{Proc. of the IEEE Conference
  on Computer Vision and Pattern Recognition}}. \bibinfo{pages}{7291--7299}.
\newblock
\urldef\tempurl%
\url{https://doi.org/10.1109/CVPR.2017.143}
\showDOI{\tempurl}


\bibitem[Carney(2005)]%
        {carney2005}
\bibfield{author}{\bibinfo{person}{Dana~R. Carney, Dana}.}
  \bibinfo{year}{2005}\natexlab{}.
\newblock \showarticletitle{Beliefs about the nonverbal expression of social
  power}.
\newblock \bibinfo{journal}{\emph{Journal of nonverbal behavior.}}
  \bibinfo{volume}{29}, \bibinfo{number}{2} (\bibinfo{year}{2005}).
\newblock
\showISSN{0191-5886}
\showLCCN{2005233986}


\bibitem[Celiktutan et~al\mbox{.}(2019)]%
        {celiktutan_multimodal_2019}
\bibfield{author}{\bibinfo{person}{Oya Celiktutan}, \bibinfo{person}{Efstratios
  Skordos}, {and} \bibinfo{person}{Hatice Gunes}.}
  \bibinfo{year}{2019}\natexlab{}.
\newblock \showarticletitle{Multimodal {Human}-{Human}-{Robot} {Interactions}
  ({MHHRI}) {Dataset} for {Studying} {Personality} and {Engagement}}.
\newblock \bibinfo{journal}{\emph{IEEE Transactions on Affective Computing}}
  \bibinfo{volume}{10}, \bibinfo{number}{4} (\bibinfo{year}{2019}),
  \bibinfo{pages}{484--497}.
\newblock
\showISSN{1949-3045}
\urldef\tempurl%
\url{https://doi.org/10.1109/TAFFC.2017.2737019}
\showDOI{\tempurl}


\bibitem[Contributors(2020)]%
        {2020mmaction2}
\bibfield{author}{\bibinfo{person}{MMAction2 Contributors}.}
  \bibinfo{year}{2020}\natexlab{}.
\newblock \bibinfo{title}{OpenMMLab's Next Generation Video Understanding
  Toolbox and Benchmark}.
\newblock
  \bibinfo{howpublished}{\url{https://github.com/open-mmlab/mmaction2}}.
\newblock


\bibitem[Engwall and Lopes(2020)]%
        {engwall2020interaction}
\bibfield{author}{\bibinfo{person}{Olov Engwall} {and}
  \bibinfo{person}{Jos{\'e} Lopes}.} \bibinfo{year}{2020}\natexlab{}.
\newblock \showarticletitle{Interaction and collaboration in robot-assisted
  language learning for adults}.
\newblock \bibinfo{journal}{\emph{Computer Assisted Language Learning}}
  (\bibinfo{year}{2020}), \bibinfo{pages}{1273--1309}.
\newblock
\urldef\tempurl%
\url{https://doi.org/10.1080/09588221.2020.1799821}
\showDOI{\tempurl}


\bibitem[Eyben et~al\mbox{.}(2015)]%
        {eyben2015geneva}
\bibfield{author}{\bibinfo{person}{Florian Eyben}, \bibinfo{person}{Klaus~R
  Scherer}, \bibinfo{person}{Bj{\"o}rn~W Schuller}, \bibinfo{person}{Johan
  Sundberg}, \bibinfo{person}{Elisabeth Andr{\'e}}, \bibinfo{person}{Carlos
  Busso}, \bibinfo{person}{Laurence~Y Devillers}, \bibinfo{person}{Julien
  Epps}, \bibinfo{person}{Petri Laukka}, \bibinfo{person}{Shrikanth~S
  Narayanan}, {et~al\mbox{.}}} \bibinfo{year}{2015}\natexlab{}.
\newblock \showarticletitle{The Geneva minimalistic acoustic parameter set
  (GeMAPS) for voice research and affective computing}.
\newblock \bibinfo{journal}{\emph{IEEE Transactions on Affective Computing}}
  \bibinfo{volume}{7}, \bibinfo{number}{2} (\bibinfo{year}{2015}),
  \bibinfo{pages}{190--202}.
\newblock
\urldef\tempurl%
\url{https://doi.org/10.1109/TAFFC.2015.2457417}
\showDOI{\tempurl}


\bibitem[Fu and Ngai(2021)]%
        {fu2021using}
\bibfield{author}{\bibinfo{person}{Eugene~Yujun Fu} {and}
  \bibinfo{person}{Michael~W Ngai}.} \bibinfo{year}{2021}\natexlab{}.
\newblock \showarticletitle{Using Motion Histories for Eye Contact Detection in
  Multiperson Group Conversations}. In \bibinfo{booktitle}{\emph{Proc. of the
  ACM International Conference on Multimedia}}. \bibinfo{pages}{4873--4877}.
\newblock
\urldef\tempurl%
\url{https://doi.org/10.1145/3474085.3479230}
\showDOI{\tempurl}


\bibitem[Goldberg et~al\mbox{.}(2021)]%
        {goldberg_attentive_2021}
\bibfield{author}{\bibinfo{person}{Patricia Goldberg}, \bibinfo{person}{Ömer
  Sümer}, \bibinfo{person}{Kathleen Stürmer}, \bibinfo{person}{Wolfgang
  Wagner}, \bibinfo{person}{Richard Göllner}, \bibinfo{person}{Peter Gerjets},
  \bibinfo{person}{Enkelejda Kasneci}, {and} \bibinfo{person}{Ulrich
  Trautwein}.} \bibinfo{year}{2021}\natexlab{}.
\newblock \showarticletitle{Attentive or {Not}? {Toward} a {Machine} {Learning}
  {Approach} to {Assessing} {Students}’ {Visible} {Engagement} in {Classroom}
  {Instruction}}.
\newblock \bibinfo{journal}{\emph{Educational Psychology Review}}
  \bibinfo{volume}{33}, \bibinfo{number}{1} (\bibinfo{year}{2021}),
  \bibinfo{pages}{27--49}.
\newblock
\showISSN{1573-336X}
\urldef\tempurl%
\url{https://doi.org/10.1007/s10648-019-09514-z}
\showDOI{\tempurl}


\bibitem[Goswami et~al\mbox{.}(2020)]%
        {goswami_towards_2020}
\bibfield{author}{\bibinfo{person}{Mononito Goswami}, \bibinfo{person}{Minkush
  Manuja}, {and} \bibinfo{person}{Maitree Leekha}.}
  \bibinfo{year}{2020}\natexlab{}.
\newblock \showarticletitle{Towards social \& engaging peer learning:
  Predicting backchanneling and disengagement in children}.
\newblock \bibinfo{journal}{\emph{arXiv preprint arXiv:2007.11346}}
  (\bibinfo{year}{2020}).
\newblock


\bibitem[Guhan et~al\mbox{.}(2020)]%
        {guhan_met_2022}
\bibfield{author}{\bibinfo{person}{Pooja Guhan}, \bibinfo{person}{Naman
  Awasthi}, \bibinfo{person}{Kristin Bussell}, \bibinfo{person}{Dinesh
  Manocha}, \bibinfo{person}{Gloria Reeves}, \bibinfo{person}{Aniket Bera},
  {et~al\mbox{.}}} \bibinfo{year}{2020}\natexlab{}.
\newblock \showarticletitle{Developing an Effective and Automated Patient
  Engagement Estimator for Telehealth: A Machine Learning Approach}.
\newblock \bibinfo{journal}{\emph{arXiv preprint arXiv:2011.08690}}
  (\bibinfo{year}{2020}).
\newblock


\bibitem[Hall et~al\mbox{.}(2005)]%
        {hall2005}
\bibfield{author}{\bibinfo{person}{Judith Hall}, \bibinfo{person}{Erik Coats},
  {and} \bibinfo{person}{Lavonia LeBeau}.} \bibinfo{year}{2005}\natexlab{}.
\newblock \showarticletitle{Nonverbal Behavior and the Vertical Dimension of
  Social Relations: A Meta-Analysis.}
\newblock \bibinfo{journal}{\emph{Psychological bulletin}}
  \bibinfo{volume}{131} (\bibinfo{date}{12} \bibinfo{year}{2005}),
  \bibinfo{pages}{898--924}.
\newblock
\urldef\tempurl%
\url{https://doi.org/10.1037/0033-2909.131.6.898}
\showDOI{\tempurl}


\bibitem[Hradis et~al\mbox{.}(2012)]%
        {hradis_voice_2012}
\bibfield{author}{\bibinfo{person}{Michal Hradis}, \bibinfo{person}{Shahram
  Eivazi}, {and} \bibinfo{person}{Roman Bednarik}.}
  \bibinfo{year}{2012}\natexlab{}.
\newblock \showarticletitle{Voice activity detection from gaze in video
  mediated communication}. In \bibinfo{booktitle}{\emph{Proc. of the ACM
  {Symposium} on {Eye} {Tracking} {Research} and {Applications}}}.
  \bibinfo{pages}{329--332}.
\newblock
\urldef\tempurl%
\url{https://doi.org/10.1145/2168556.2168628}
\showDOI{\tempurl}


\bibitem[Jain et~al\mbox{.}(2020)]%
        {jain_modeling_2020}
\bibfield{author}{\bibinfo{person}{Shomik Jain},
  \bibinfo{person}{Balasubramanian Thiagarajan}, \bibinfo{person}{Zhonghao
  Shi}, \bibinfo{person}{Caitlyn Clabaugh}, {and} \bibinfo{person}{Maja~J
  Matari{\'c}}.} \bibinfo{year}{2020}\natexlab{}.
\newblock \showarticletitle{Modeling engagement in long-term, in-home socially
  assistive robot interventions for children with autism spectrum disorders}.
\newblock \bibinfo{journal}{\emph{Science Robotics}} \bibinfo{volume}{5},
  \bibinfo{number}{39} (\bibinfo{year}{2020}).
\newblock
\urldef\tempurl%
\url{https://doi.org/10.1126/scirobotics.aaz3791}
\showDOI{\tempurl}


\bibitem[Karimah and Hasegawa(2021)]%
        {karimah_automatic_2021}
\bibfield{author}{\bibinfo{person}{Shofiyati~Nur Karimah} {and}
  \bibinfo{person}{Shinobu Hasegawa}.} \bibinfo{year}{2021}\natexlab{}.
\newblock \showarticletitle{Automatic {Engagement} {Recognition} for {Distance}
  {Learning} {Systems}: {A} {Literature} {Study} of {Engagement} {Datasets} and
  {Methods}}. In \bibinfo{booktitle}{\emph{Augmented {Cognition}}}
  \emph{(\bibinfo{series}{Lecture {Notes} in {Computer} {Science}})}.
  \bibinfo{publisher}{Springer International Publishing},
  \bibinfo{address}{Cham}, \bibinfo{pages}{264--276}.
\newblock
\showISBNx{978-3-030-78114-9}
\urldef\tempurl%
\url{https://doi.org/10.1007/978-3-030-78114-9_19}
\showDOI{\tempurl}


\bibitem[Lin(1989)]%
        {ccc}
\bibfield{author}{\bibinfo{person}{Lawrence I-Kuei Lin}.}
  \bibinfo{year}{1989}\natexlab{}.
\newblock \showarticletitle{A Concordance Correlation Coefficient to Evaluate
  Reproducibility}.
\newblock \bibinfo{journal}{\emph{Biometrics}} \bibinfo{volume}{45},
  \bibinfo{number}{1} (\bibinfo{year}{1989}), \bibinfo{pages}{255--268}.
\newblock
\showISSN{0006341X, 15410420}
\urldef\tempurl%
\url{https://doi.org/10.2307/2532051}
\showDOI{\tempurl}


\bibitem[Liu et~al\mbox{.}(2021b)]%
        {liu2021imigue}
\bibfield{author}{\bibinfo{person}{Xin Liu}, \bibinfo{person}{Henglin Shi},
  \bibinfo{person}{Haoyu Chen}, \bibinfo{person}{Zitong Yu},
  \bibinfo{person}{Xiaobai Li}, {and} \bibinfo{person}{Guoying Zhao}.}
  \bibinfo{year}{2021}\natexlab{b}.
\newblock \showarticletitle{iMiGUE: An identity-free video dataset for
  micro-gesture understanding and emotion analysis}. In
  \bibinfo{booktitle}{\emph{Proc. of the IEEE Conference on Computer Vision and
  Pattern Recognition}}. \bibinfo{pages}{10631--10642}.
\newblock
\urldef\tempurl%
\url{https://doi.org/10.1109/CVPR46437.2021.01049}
\showDOI{\tempurl}


\bibitem[Liu et~al\mbox{.}(2021a)]%
        {liu2021Swin}
\bibfield{author}{\bibinfo{person}{Ze Liu}, \bibinfo{person}{Yutong Lin},
  \bibinfo{person}{Yue Cao}, \bibinfo{person}{Han Hu}, \bibinfo{person}{Yixuan
  Wei}, \bibinfo{person}{Zheng Zhang}, \bibinfo{person}{Stephen Lin}, {and}
  \bibinfo{person}{Baining Guo}.} \bibinfo{year}{2021}\natexlab{a}.
\newblock \showarticletitle{Swin Transformer: Hierarchical Vision Transformer
  using Shifted Windows}. In \bibinfo{booktitle}{\emph{Proc. of the IEEE/CVF
  International Conference on Computer Vision}}. \bibinfo{pages}{10012--10022}.
\newblock
\urldef\tempurl%
\url{https://doi.org/10.1109/ICCV48922.2021.00986}
\showDOI{\tempurl}


\bibitem[Lopes et~al\mbox{.}(2017)]%
        {lopes2017first}
\bibfield{author}{\bibinfo{person}{Jos{\'e} Lopes}, \bibinfo{person}{Olov
  Engwall}, {and} \bibinfo{person}{Gabriel Skantze}.}
  \bibinfo{year}{2017}\natexlab{}.
\newblock \showarticletitle{A first visit to the robot language caf{\'e}}. In
  \bibinfo{booktitle}{\emph{Proc. of the ISCA Workshop on Speech and Language
  Technology in Education}}.
\newblock
\urldef\tempurl%
\url{https://doi.org/10.21437/SLaTE.2017-2}
\showDOI{\tempurl}


\bibitem[Ma et~al\mbox{.}(2022)]%
        {ma2022ta}
\bibfield{author}{\bibinfo{person}{Fuyan Ma}, \bibinfo{person}{Ziyu Ma},
  \bibinfo{person}{Bin Sun}, {and} \bibinfo{person}{Shutao Li}.}
  \bibinfo{year}{2022}\natexlab{}.
\newblock \showarticletitle{TA-CNN: A Unified Network for Human Behavior
  Analysis in Multi-Person Conversations}. In \bibinfo{booktitle}{\emph{Proc.
  of the ACM International Conference on Multimedia}}.
  \bibinfo{pages}{7099--7103}.
\newblock
\urldef\tempurl%
\url{https://doi.org/10.1145/3503161.3551587}
\showDOI{\tempurl}


\bibitem[Mehrabian(1968)]%
        {mehrabian1968relationship}
\bibfield{author}{\bibinfo{person}{Albert Mehrabian}.}
  \bibinfo{year}{1968}\natexlab{}.
\newblock \showarticletitle{Relationship of attitude to seated posture,
  orientation, and distance.}
\newblock \bibinfo{journal}{\emph{Journal of personality and social
  psychology}} \bibinfo{volume}{10}, \bibinfo{number}{1}
  (\bibinfo{year}{1968}), \bibinfo{pages}{26}.
\newblock
\urldef\tempurl%
\url{https://doi.org/10.1037/h0026384}
\showDOI{\tempurl}


\bibitem[Mehrabian and Friar(1969)]%
        {mehrabian1969encoding}
\bibfield{author}{\bibinfo{person}{Albert Mehrabian} {and}
  \bibinfo{person}{John~T Friar}.} \bibinfo{year}{1969}\natexlab{}.
\newblock \showarticletitle{Encoding of attitude by a seated communicator via
  posture and position cues.}
\newblock \bibinfo{journal}{\emph{Journal of Consulting and Clinical
  Psychology}} \bibinfo{volume}{33}, \bibinfo{number}{3}
  (\bibinfo{year}{1969}), \bibinfo{pages}{330}.
\newblock
\urldef\tempurl%
\url{https://doi.org/10.1037/h0027576}
\showDOI{\tempurl}


\bibitem[Mohiyeddini et~al\mbox{.}(2013)]%
        {mohiyeddini2013displacement}
\bibfield{author}{\bibinfo{person}{Changiz Mohiyeddini},
  \bibinfo{person}{Stephanie Bauer}, {and} \bibinfo{person}{Stuart Semple}.}
  \bibinfo{year}{2013}\natexlab{}.
\newblock \showarticletitle{Displacement behaviour is associated with reduced
  stress levels among men but not women}.
\newblock \bibinfo{journal}{\emph{PloS one}} \bibinfo{volume}{8},
  \bibinfo{number}{2} (\bibinfo{year}{2013}), \bibinfo{pages}{e56355}.
\newblock
\urldef\tempurl%
\url{https://doi.org/10.1371/journal.pone.0056355}
\showDOI{\tempurl}


\bibitem[Mohiyeddini et~al\mbox{.}(2015)]%
        {mohiyeddini2015neuroticism}
\bibfield{author}{\bibinfo{person}{Changiz Mohiyeddini},
  \bibinfo{person}{Stephanie Bauer}, {and} \bibinfo{person}{Stuart Semple}.}
  \bibinfo{year}{2015}\natexlab{}.
\newblock \showarticletitle{Neuroticism and stress: The role of displacement
  behavior}.
\newblock \bibinfo{journal}{\emph{Anxiety, stress, \& coping}}
  \bibinfo{volume}{28}, \bibinfo{number}{4} (\bibinfo{year}{2015}),
  \bibinfo{pages}{391--407}.
\newblock
\urldef\tempurl%
\url{https://doi.org/10.1080/10615806.2014.1000878}
\showDOI{\tempurl}


\bibitem[M\"uller and Bulling(2019)]%
        {muller2019emergent}
\bibfield{author}{\bibinfo{person}{Philipp M\"uller} {and}
  \bibinfo{person}{Andreas Bulling}.} \bibinfo{year}{2019}\natexlab{}.
\newblock \showarticletitle{Emergent Leadership Detection Across Datasets}. In
  \bibinfo{booktitle}{\emph{Proc. of the ACM International Conference on
  Multimodal Interaction}}. \bibinfo{pages}{274--278}.
\newblock
\urldef\tempurl%
\url{https://doi.org/10.1145/3340555.3353721}
\showDOI{\tempurl}


\bibitem[M{\"u}ller et~al\mbox{.}(2022)]%
        {muller2022multimediate}
\bibfield{author}{\bibinfo{person}{Philipp M{\"u}ller},
  \bibinfo{person}{Michael Dietz}, \bibinfo{person}{Dominik Schiller},
  \bibinfo{person}{Dominike Thomas}, \bibinfo{person}{Hali Lindsay},
  \bibinfo{person}{Patrick Gebhard}, \bibinfo{person}{Elisabeth Andr{\'e}},
  {and} \bibinfo{person}{Andreas Bulling}.} \bibinfo{year}{2022}\natexlab{}.
\newblock \showarticletitle{MultiMediate'22: Backchannel Detection and
  Agreement Estimation in Group Interactions}. In
  \bibinfo{booktitle}{\emph{Proc. of the ACM International Conference on
  Multimedia}}. \bibinfo{pages}{7109--7114}.
\newblock
\urldef\tempurl%
\url{https://doi.org/10.1145/3503161.3551589}
\showDOI{\tempurl}


\bibitem[M{\"u}ller et~al\mbox{.}(2021)]%
        {muller2021multimediate}
\bibfield{author}{\bibinfo{person}{Philipp M{\"u}ller},
  \bibinfo{person}{Michael Dietz}, \bibinfo{person}{Dominik Schiller},
  \bibinfo{person}{Dominike Thomas}, \bibinfo{person}{Guanhua Zhang},
  \bibinfo{person}{Patrick Gebhard}, \bibinfo{person}{Elisabeth Andr{\'e}},
  {and} \bibinfo{person}{Andreas Bulling}.} \bibinfo{year}{2021}\natexlab{}.
\newblock \showarticletitle{MultiMediate: Multi-modal Group Behaviour Analysis
  for Artificial Mediation}. In \bibinfo{booktitle}{\emph{Proc. of the ACM
  International Conference on Multimedia}}. \bibinfo{pages}{4878--4882}.
\newblock
\urldef\tempurl%
\url{https://doi.org/10.1145/3474085.3479219}
\showDOI{\tempurl}


\bibitem[M{\"u}ller et~al\mbox{.}(2018)]%
        {muller2018robust}
\bibfield{author}{\bibinfo{person}{Philipp M{\"u}ller},
  \bibinfo{person}{Michael~Xuelin Huang}, \bibinfo{person}{Xucong Zhang}, {and}
  \bibinfo{person}{Andreas Bulling}.} \bibinfo{year}{2018}\natexlab{}.
\newblock \showarticletitle{Robust eye contact detection in natural
  multi-person interactions using gaze and speaking behaviour}. In
  \bibinfo{booktitle}{\emph{Proc. of the ACM Symposium on Eye Tracking Research
  \& Applications}}. \bibinfo{pages}{1--10}.
\newblock
\urldef\tempurl%
\url{https://doi.org/10.1145/3204493.3204549}
\showDOI{\tempurl}


\bibitem[Müller et~al\mbox{.}(2018)]%
        {muller_detecting_2018}
\bibfield{author}{\bibinfo{person}{Philipp Müller},
  \bibinfo{person}{Michael~Xuelin Huang}, {and} \bibinfo{person}{Andreas
  Bulling}.} \bibinfo{year}{2018}\natexlab{}.
\newblock \showarticletitle{Detecting {Low} {Rapport} {During} {Natural}
  {Interactions} in {Small} {Groups} from {Non}-{Verbal} {Behaviour}}. In
  \bibinfo{booktitle}{\emph{Proc. of the ACM {International} {Conference} on
  {Intelligent} {User} {Interfaces}}}. \bibinfo{publisher}{Association for
  Computing Machinery}, \bibinfo{pages}{153--164}.
\newblock
\showISBNx{978-1-4503-4945-1}
\urldef\tempurl%
\url{https://doi.org/10.1145/3172944.3172969}
\showDOI{\tempurl}


\bibitem[Oertel et~al\mbox{.}(2020)]%
        {oertel_engagement_2020}
\bibfield{author}{\bibinfo{person}{Catharine Oertel}, \bibinfo{person}{Ginevra
  Castellano}, \bibinfo{person}{Mohamed Chetouani}, \bibinfo{person}{Jauwairia
  Nasir}, \bibinfo{person}{Mohammad Obaid}, \bibinfo{person}{Catherine
  Pelachaud}, {and} \bibinfo{person}{Christopher Peters}.}
  \bibinfo{year}{2020}\natexlab{}.
\newblock \showarticletitle{Engagement in {Human}-{Agent} {Interaction}: {An}
  {Overview}}.
\newblock \bibinfo{journal}{\emph{Frontiers in Robotics and AI}}
  \bibinfo{volume}{7} (\bibinfo{year}{2020}).
\newblock
\showISSN{2296-9144}
\urldef\tempurl%
\url{https://doi.org/10.3389/frobt.2020.00092}
\showDOI{\tempurl}


\bibitem[Oertel and Salvi(2013)]%
        {oertel_gaze-based_2013}
\bibfield{author}{\bibinfo{person}{Catharine Oertel} {and}
  \bibinfo{person}{Giampiero Salvi}.} \bibinfo{year}{2013}\natexlab{}.
\newblock \showarticletitle{A gaze-based method for relating group involvement
  to individual engagement in multimodal multiparty dialogue}. In
  \bibinfo{booktitle}{\emph{Proc. of the ACM International Conference on
  Multimodal Interaction}}. \bibinfo{pages}{99--106}.
\newblock
\urldef\tempurl%
\url{https://doi.org/10.1145/2522848.2522865}
\showDOI{\tempurl}


\bibitem[Ohshima et~al\mbox{.}(2017)]%
        {ohshima_neut_2017}
\bibfield{author}{\bibinfo{person}{N. Ohshima}, \bibinfo{person}{R. Fujimori},
  \bibinfo{person}{H. Tokunaga}, \bibinfo{person}{H. Kaneko}, {and}
  \bibinfo{person}{N. Mukawa}.} \bibinfo{year}{2017}\natexlab{}.
\newblock \showarticletitle{Neut: {Design} and evaluation of speaker
  designation behaviors for communication support robot to encourage
  conversations}. In \bibinfo{booktitle}{\emph{Proc. of the {IEEE}
  {International} {Symposium} on {Robot} and {Human} {Interactive}
  {Communication} ({RO}-{MAN})}}. \bibinfo{pages}{1387--1393}.
\newblock
\urldef\tempurl%
\url{https://doi.org/10.1109/ROMAN.2017.8172485}
\showDOI{\tempurl}


\bibitem[O'Malley et~al\mbox{.}(2019)]%
        {omalley2019kerastuner}
\bibfield{author}{\bibinfo{person}{Tom O'Malley}, \bibinfo{person}{Elie
  Bursztein}, \bibinfo{person}{James Long}, \bibinfo{person}{Fran\c{c}ois
  Chollet}, \bibinfo{person}{Haifeng Jin}, \bibinfo{person}{Luca Invernizzi},
  {et~al\mbox{.}}} \bibinfo{year}{2019}\natexlab{}.
\newblock \bibinfo{title}{KerasTuner}.
\newblock
  \bibinfo{howpublished}{\url{https://github.com/keras-team/keras-tuner}}.
\newblock


\bibitem[Park et~al\mbox{.}(2019)]%
        {park_model-free_2019}
\bibfield{author}{\bibinfo{person}{Hae~Won Park}, \bibinfo{person}{Ishaan
  Grover}, \bibinfo{person}{Samuel Spaulding}, \bibinfo{person}{Louis Gomez},
  {and} \bibinfo{person}{Cynthia Breazeal}.} \bibinfo{year}{2019}\natexlab{}.
\newblock \showarticletitle{A {Model}-{Free} {Affective} {Reinforcement}
  {Learning} {Approach} to {Personalization} of an {Autonomous} {Social}
  {Robot} {Companion} for {Early} {Literacy} {Education}}. In
  \bibinfo{booktitle}{\emph{Proc. of the AAAI Conference on Artificial
  Intelligence}}. \bibinfo{pages}{687--694}.
\newblock
\urldef\tempurl%
\url{https://doi.org/10.1609/aaai.v33i01.3301687}
\showDOI{\tempurl}


\bibitem[Park and Lim(2020)]%
        {park2020investigating}
\bibfield{author}{\bibinfo{person}{Sunjeong Park} {and}
  \bibinfo{person}{Youn-kyung Lim}.} \bibinfo{year}{2020}\natexlab{}.
\newblock \showarticletitle{Investigating User Expectations on the Roles of
  Family-shared AI Speakers}. In \bibinfo{booktitle}{\emph{Proc. of the ACM
  Conference on Human Factors in Computing Systems}}. \bibinfo{pages}{1--13}.
\newblock
\urldef\tempurl%
\url{https://doi.org/10.1145/3313831.3376450}
\showDOI{\tempurl}


\bibitem[Peters et~al\mbox{.}(2005)]%
        {peters_engagement_2005}
\bibfield{author}{\bibinfo{person}{Christopher Peters},
  \bibinfo{person}{Catherine Pelachaud}, \bibinfo{person}{Elisabetta Bevacqua},
  \bibinfo{person}{Maurizio Mancini}, {and} \bibinfo{person}{Isabella Poggi}.}
  \bibinfo{year}{2005}\natexlab{}.
\newblock \showarticletitle{Engagement {Capabilities} for {ECAs}}.
\newblock \bibinfo{journal}{\emph{Autonomous Agents and Multi-agent Systems -
  AAMAS}} (\bibinfo{year}{2005}).
\newblock


\bibitem[Rajagopalan et~al\mbox{.}(2015)]%
        {rajagopalan_play_2015}
\bibfield{author}{\bibinfo{person}{Shyam~Sundar Rajagopalan},
  \bibinfo{person}{O.V.~Ramana Murthy}, \bibinfo{person}{Roland Goecke}, {and}
  \bibinfo{person}{Agata Rozga}.} \bibinfo{year}{2015}\natexlab{}.
\newblock \showarticletitle{Play with me — {Measuring} a child's engagement
  in a social interaction}. In \bibinfo{booktitle}{\emph{Proc. of the {IEEE}
  {International} {Conference} and {Workshops} on {Automatic} {Face} and
  {Gesture} {Recognition}}}, Vol.~\bibinfo{volume}{1}.
\newblock
\urldef\tempurl%
\url{https://doi.org/10.1109/FG.2015.7163129}
\showDOI{\tempurl}


\bibitem[Rich et~al\mbox{.}(2010)]%
        {CharlesRichEngagementHumanRobotInteraction}
\bibfield{author}{\bibinfo{person}{Charles Rich}, \bibinfo{person}{Brett
  Ponsler}, \bibinfo{person}{Aaron Holroyd}, {and} \bibinfo{person}{Candace~L
  Sidner}.} \bibinfo{year}{2010}\natexlab{}.
\newblock \showarticletitle{Recognizing engagement in human-robot interaction}.
  In \bibinfo{booktitle}{\emph{Proc. of the ACM/IEEE International Conference
  on Human-Robot Interaction}}. IEEE, \bibinfo{pages}{375--382}.
\newblock
\urldef\tempurl%
\url{https://doi.org/10.1109/HRI.2010.5453163}
\showDOI{\tempurl}


\bibitem[Ringeval et~al\mbox{.}(2013)]%
        {ringeval_introducing_2013}
\bibfield{author}{\bibinfo{person}{Fabien Ringeval}, \bibinfo{person}{Andreas
  Sonderegger}, \bibinfo{person}{Juergen Sauer}, {and} \bibinfo{person}{Denis
  Lalanne}.} \bibinfo{year}{2013}\natexlab{}.
\newblock \showarticletitle{Introducing the {RECOLA} multimodal corpus of
  remote collaborative and affective interactions}. In
  \bibinfo{booktitle}{\emph{Proc. of the {IEEE} {International} {Conference}
  and {Workshops} on {Automatic} {Face} and {Gesture} {Recognition} ({FG})}}.
\newblock
\urldef\tempurl%
\url{https://doi.org/10.1109/FG.2013.6553805}
\showDOI{\tempurl}


\bibitem[Sanghvi et~al\mbox{.}(2011)]%
        {Sanghvi:2011:AAA:1957656.1957781BodyMotion}
\bibfield{author}{\bibinfo{person}{Jyotirmay Sanghvi}, \bibinfo{person}{Ginevra
  Castellano}, \bibinfo{person}{Iolanda Leite}, \bibinfo{person}{Andr{\'e}
  Pereira}, \bibinfo{person}{Peter~W. McOwan}, {and} \bibinfo{person}{Ana
  Paiva}.} \bibinfo{year}{2011}\natexlab{}.
\newblock \showarticletitle{Automatic Analysis of Affective Postures and Body
  Motion to Detect Engagement with a Game Companion}. In
  \bibinfo{booktitle}{\emph{Proc. of the ACM/IEEE International Conference on
  Human-robot Interaction}}. \bibinfo{pages}{305--312}.
\newblock
\showISBNx{978-1-4503-0561-7}


\bibitem[Sebo et~al\mbox{.}(2020)]%
        {sebo2020robots}
\bibfield{author}{\bibinfo{person}{Sarah Sebo}, \bibinfo{person}{Brett Stoll},
  \bibinfo{person}{Brian Scassellati}, {and} \bibinfo{person}{Malte~F Jung}.}
  \bibinfo{year}{2020}\natexlab{}.
\newblock \showarticletitle{Robots in groups and teams: a literature review}.
\newblock \bibinfo{journal}{\emph{Proc. of the ACM on Human-Computer
  Interaction}} \bibinfo{volume}{4}, \bibinfo{number}{CSCW2}
  (\bibinfo{year}{2020}), \bibinfo{pages}{1--36}.
\newblock
\urldef\tempurl%
\url{https://doi.org/10.1145/3415247}
\showDOI{\tempurl}


\bibitem[Sharma et~al\mbox{.}(2022)]%
        {sharma2022graph}
\bibfield{author}{\bibinfo{person}{Garima Sharma}, \bibinfo{person}{Kalin
  Stefanov}, \bibinfo{person}{Abhinav Dhall}, {and} \bibinfo{person}{Jianfei
  Cai}.} \bibinfo{year}{2022}\natexlab{}.
\newblock \showarticletitle{Graph-based Group Modelling for Backchannel
  Detection}. In \bibinfo{booktitle}{\emph{Proc. of the ACM International
  Conference on Multimedia}}. \bibinfo{pages}{7190--7194}.
\newblock
\urldef\tempurl%
\url{https://doi.org/10.1145/3503161.3551605}
\showDOI{\tempurl}


\bibitem[Sharpley and Sagris(1995)]%
        {sharpley1995}
\bibfield{author}{\bibinfo{person}{Christopher~F. Sharpley} {and}
  \bibinfo{person}{Anastasia Sagris}.} \bibinfo{year}{1995}\natexlab{}.
\newblock \showarticletitle{When does counsellor forward lean influence
  client-perceived rapport?}
\newblock \bibinfo{journal}{\emph{British Journal of Guidance \& Counselling}}
  \bibinfo{volume}{23}, \bibinfo{number}{3} (\bibinfo{year}{1995}),
  \bibinfo{pages}{387--394}.
\newblock
\urldef\tempurl%
\url{https://doi.org/10.1080/03069889508253696}
\showDOI{\tempurl}


\bibitem[Short and Mataric(2017)]%
        {short2017robot}
\bibfield{author}{\bibinfo{person}{Elaine Short} {and} \bibinfo{person}{Maja~J.
  Mataric}.} \bibinfo{year}{2017}\natexlab{}.
\newblock \showarticletitle{Robot moderation of a collaborative game: Towards
  socially assistive robotics in group interactions}. In
  \bibinfo{booktitle}{\emph{Proc. of the IEEE International Symposium on Robot
  and Human Interactive Communication}}. \bibinfo{pages}{385--390}.
\newblock
\urldef\tempurl%
\url{https://doi.org/10.1109/ROMAN.2017.8172331}
\showDOI{\tempurl}


\bibitem[Simon et~al\mbox{.}(2017)]%
        {simon2017hand}
\bibfield{author}{\bibinfo{person}{Tomas Simon}, \bibinfo{person}{Hanbyul Joo},
  \bibinfo{person}{Iain Matthews}, {and} \bibinfo{person}{Yaser Sheikh}.}
  \bibinfo{year}{2017}\natexlab{}.
\newblock \showarticletitle{Hand keypoint detection in single images using
  multiview bootstrapping}. In \bibinfo{booktitle}{\emph{Proc. of the IEEE
  Conference on Computer Vision and Pattern Recognition}}.
  \bibinfo{pages}{1145--1153}.
\newblock
\urldef\tempurl%
\url{https://doi.org/10.1109/CVPR.2017.494}
\showDOI{\tempurl}


\bibitem[Utami and Bickmore(2019)]%
        {utami2019collaborative}
\bibfield{author}{\bibinfo{person}{Dina Utami} {and} \bibinfo{person}{Timothy
  Bickmore}.} \bibinfo{year}{2019}\natexlab{}.
\newblock \showarticletitle{Collaborative user responses in multiparty
  interaction with a couples counselor robot}. In
  \bibinfo{booktitle}{\emph{Proc. of the ACM/IEEE International Conference on
  Human-Robot Interaction}}. \bibinfo{pages}{294--303}.
\newblock
\urldef\tempurl%
\url{https://doi.org/10.1109/HRI.2019.8673177}
\showDOI{\tempurl}


\bibitem[Vacharkulksemsuk et~al\mbox{.}(2016)]%
        {vacharkulksemsuk2016dominant}
\bibfield{author}{\bibinfo{person}{Tanya Vacharkulksemsuk},
  \bibinfo{person}{Emily Reit}, \bibinfo{person}{Poruz Khambatta},
  \bibinfo{person}{Paul~W Eastwick}, \bibinfo{person}{Eli~J Finkel}, {and}
  \bibinfo{person}{Dana~R Carney}.} \bibinfo{year}{2016}\natexlab{}.
\newblock \showarticletitle{Dominant, open nonverbal displays are attractive at
  zero-acquaintance}.
\newblock \bibinfo{journal}{\emph{Proceedings of the National Academy of
  Sciences}} \bibinfo{volume}{113}, \bibinfo{number}{15}
  (\bibinfo{year}{2016}), \bibinfo{pages}{4009--4014}.
\newblock
\urldef\tempurl%
\url{https://doi.org/10.1073/pnas.1508932113}
\showDOI{\tempurl}


\bibitem[Valstar et~al\mbox{.}(2016)]%
        {valstar2016avec}
\bibfield{author}{\bibinfo{person}{Michel Valstar}, \bibinfo{person}{Jonathan
  Gratch}, \bibinfo{person}{Bj{\"o}rn Schuller}, \bibinfo{person}{Fabien
  Ringeval}, \bibinfo{person}{Denis Lalanne}, \bibinfo{person}{Mercedes
  Torres~Torres}, \bibinfo{person}{Stefan Scherer}, \bibinfo{person}{Giota
  Stratou}, \bibinfo{person}{Roddy Cowie}, {and} \bibinfo{person}{Maja
  Pantic}.} \bibinfo{year}{2016}\natexlab{}.
\newblock \showarticletitle{Avec 2016: Depression, mood, and emotion
  recognition workshop and challenge}. In \bibinfo{booktitle}{\emph{Proc. of
  the International Workshop on Audio/Visual Emotion Challenge}}.
  \bibinfo{pages}{3--10}.
\newblock
\urldef\tempurl%
\url{https://doi.org/10.1145/2988257.2988258}
\showDOI{\tempurl}


\bibitem[Wagner et~al\mbox{.}(2018)]%
        {DBLP:conf/interspeech/0001SSA18}
\bibfield{author}{\bibinfo{person}{Johannes Wagner}, \bibinfo{person}{Dominik
  Schiller}, \bibinfo{person}{Andreas Seiderer}, {and}
  \bibinfo{person}{Elisabeth Andr{\'{e}}}.} \bibinfo{year}{2018}\natexlab{}.
\newblock \showarticletitle{Deep Learning in Paralinguistic Recognition Tasks:
  Are Hand-crafted Features Still Relevant?}. In
  \bibinfo{booktitle}{\emph{Proc. Interspeech}},
  \bibfield{editor}{\bibinfo{person}{B.~Yegnanarayana}} (Ed.).
  \bibinfo{pages}{147--151}.
\newblock
\urldef\tempurl%
\url{https://doi.org/10.21437/Interspeech.2018-1238}
\showDOI{\tempurl}


\bibitem[Wallbott(1998)]%
        {wallbott1998}
\bibfield{author}{\bibinfo{person}{Harald~G Wallbott}.}
  \bibinfo{year}{1998}\natexlab{}.
\newblock \showarticletitle{Bodily expression of emotion}.
\newblock \bibinfo{journal}{\emph{European journal of social psychology.}}
  \bibinfo{volume}{28}, \bibinfo{number}{6} (\bibinfo{year}{1998}).
\newblock
\showISSN{0046-2772}
\showLCCN{2001212331}


\end{thebibliography}

\end{document}